\documentclass[sigconf]{acmart}

\usepackage{booktabs} 
\usepackage{graphicx}
\usepackage{booktabs}
\usepackage{flushend}
\usepackage{cleveref}
\usepackage{subcaption}
\graphicspath{ {Images/} }

\renewcommand{\vec}[1]{\mathbf{#1}}

\acmDOI{10.1145/nnnnnnn.nnnnnnn}

\acmConference[GECCO '20]{the Genetic and Evolutionary Computation Conference 2020}{July 8--12, 2020}{Cancun, Mexico}
\acmYear{2020}
\copyrightyear{2020}

\acmPrice{15.00}

\begin{document}
\title{Diversity-based Design Assist for Large Legged Robots}

\author{David Howard}
\orcid{1234-5678-9012}
\affiliation{%
  \institution{Robotics and Autonomous Systems Group, CSIRO}
  \streetaddress{P.O. Box 1212}
  \city{Brisbane} 
  \state{Queensland, Australia} 
  \postcode{4069}
}
\email{david.howard@csiro.au}

\author{Thomas Lowe}
\affiliation{%
  \institution{Robotics and Autonomous Systems Group, CSIRO}
  \streetaddress{P.O. Box 1212}
  \city{Brisbane} 
  \state{Queensland, Australia} 
  \postcode{4069}
}
\email{thomas.lowe@csiro.au}

\author{Wade Geles}
\affiliation{%
  \institution{University of Queensland}
  \streetaddress{St Lucia}
  \city{Brisbane} 
  \state{Queensland, Australia} 
  \postcode{4072}
}


\begin{abstract}
We combine MAP-Elites and highly parallelisable simulation to explore the design space of a class of large legged robots, which stand at around 2m tall and whose design and construction is not well-studied. The simulation is modified to account for factors such as motor torque and weight, and presents a reasonable fidelity search space. A novel robot encoding allows for bio-inspired features such as legs scaling along the length of the body. The impact of three possible control generation schemes are assessed in the context of body-brain co-evolution, showing that even constrained problems benefit strongly from coupling-promoting mechanisms. A two stage process in implemented. In the first stage, a library of possible robots is generated, treating user requirements as constraints. In the second stage, the most promising robot niches are analysed and a suite of human-understandable design rules generated related to the values of their feature variables. These rules, together with the library, are then ready to be used by a (human) robot designer as a Design Assist tool.

\end{abstract}

%
%



\maketitle

\section{Introduction}

Design Assist occupies the midpoint along a spectrum of compute-based design approaches that encompasses CAD\footnote{where the human is the designer and the design is digital} at one end and Computer Automated Design, or CautoD\footnote{where the algorithm is a designer and takes no  human input, c.f., an overwhelming majority of Evolutionary Robotics works.}~\cite{ang2016smart} at the other.  Design Assist's main tenet is that the compute-intensive and bias-free (although somewhat `blind') creation and assessment of a plethora of potential solutions allows human designers to access more of the design space.  When partnered With a human's creativity, intuition, and domain knowledge, a range of designs can then be considered, regardless of how outlandish or unconventional the solution may be, for further refinement. Additionally, general rules may be discovered to simplify or characterise the space and provide further domain knowledge for the problem under consideration.

For the purposes of this paper, there are two key considerations:

(1) Design Assist holds particular promise for the automated design of items that are difficult to prototype {\em in materio}, are highly multivariate, or where the mapping of features to performance is not straightforward --- in other words where the space itself presents a barrier to design discovery.  

(2) For Design Assist to be effective, it must be able to provide the human designer with a wide range of different possibilities, or views into the underlying design space, as well as offering the potential to provide information about the space itself that they would otherwise not have access to.  Our main motivation is in providing this information to our engineers.

 \begin{figure}[t]
\begin{center}

\begin{subfigure}[b]{8.5cm}
  \centering
  \includegraphics[width=8.5cm]{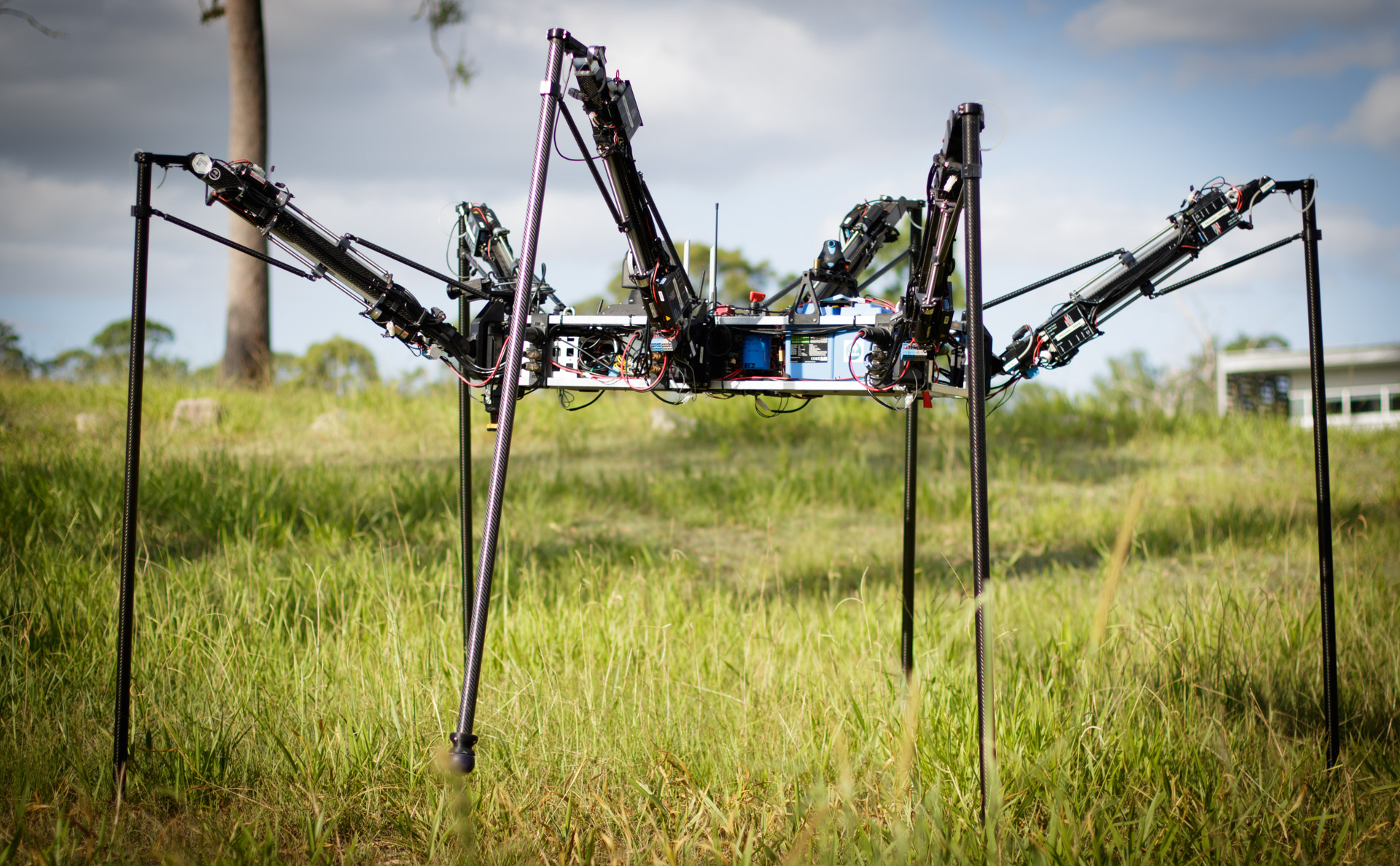}
\end{subfigure}

\end{center}
\caption{The Multilegged Autonomous eXplorer, MAX.  MAX belongs to a class of large (>2m tall) superlight (approx. 60kg) robots designed for difficult terrain traversal, and is the focus of this study.}
\label{fig1:max}
\end{figure}

Given these observations, we here report on a series of experiments that perform Design Assist on a class of large legged robots that are both costly and labour-intensive to produce.  To give some context, the first iteration of this robot, named MAX, has been previously described~\cite{elfes2017multilegged} and is shown in Fig.~\ref{fig1:max} --- the engineering team are looking for improvements to MAX for the next iteration of the platform.  

MAX is particularly interesting due to a combination of extreme size (>2m tall) and low weight (approx. 60kg). This allows it to perform missions in extreme environments e.g., rain forests, as it can step over the types of low-lying foliage that would prohibit other types of robots.  To ensure this capability is preserved, MAX's design documentation forms part of the input to the Design Assist procedure and appropriately constrain the search space.  Designers of legged robots often take inspiration from nature; this class of robot has no natural counterpart and is only possible to create through synthetic materials.  Design assist therefore presents an opportunity to find feasible regions of this under-explored space and generate design rules that can inform the design of other types of robots in this class.

It is natural for the underlying algorithm to be from the family of either Quality Diversity~\cite{pugh2016quality}, or novelty/surprise~\cite{lehman2011abandoning}, given their predisposition to generating large, varied libraries of solutions.  Our solution harnesses MAP-Elites~\cite{mouret2015illuminating} to generate a diversity of designs for their consideration, given its noted success in similar scenarios~\cite{gaier2018dataefficient}.  

Our approach is novel due to a combination of the following:

\begin{itemize}
    \item Firstly, we use a design brief to constrain the robot evolution process, and synthesise an expressive, meaningful direct encoding that captures all of the dimensions of variance we wish to consider in very few parameters.
    \item Our simulator is seeded with domain-relevant information on e.g., motor power-to-mass ratings, allowing a more true-to-life simulation.  Coupled with this, we run MAP-Elites with 6 feature dimensions to meaningfully represent the design space.
    \item After generating the `library', we analyse the best designs and feed them into a principle component analysis, to derive general design rules that capture parts of this mapping in a parsimonious way.  Given that large legged robots are not well-studied, such rules are also positioned to uncover underlying principles that can be applied to future designs within the same class of robot.
    \item Our experimental setup contributes to the expanding body of knowledge on co-evolution of robot morphology and control \cite{EvolvingRobotMorphologyAndControl}.  We contribute to this discussion by comparing three methods for body-brain coupling, with varying amounts of compute time spent on each.
\end{itemize}

The remainder of the paper is structured as follows: Section 2 deals with background and related work, and Section 3 details our methodology.  Section 4 covers the operation of the evolutionary algorithm.  Section 5 discusses the experimentation and Section 6 presents the results of our study.  Section 7 looks at how we generate design rules, and Section 8 concludes the paper.

\section{Background}

In this section we cover the two most pertinent areas of related research; Evolutionary Robotics and Diversity-based evolutionary approaches.

\subsection{Evolutionary Robotics}

Evolutionary Robotics~\cite{nolfi2016evolutionary,10.3389/frobt.2015.00004} (ER) can be broadly defined as the use of evolutionary approaches to optimise a robot's brain, body, or both, with candidates typically assessed against some user-defined fitness function related to in-environment performance.  Starting with the seminal work of Sims~\cite{EvolvingVirtualCreatures}, Evolutionary Robotics has been shown to provide creative, adaptive, and fit-for-function solutions in both simulation and, more recently, reality ~\cite{lipson2000automatic,howard2017platform}. A recent approach couples high-fidelity simulation with selectively evolution of only parts of a robot's morphology most likely to interact with the environment, whilst simultaneously benefiting from a mature, general-purpose control stack~\cite{collins2018towards}.

Legged robots are a popular candidate for evolution, with work ongoing and dating back to the 1990s.  Typically, controllers are optimised for a fixed-morphology \cite{hornby1999autonomous,chernova2004evolutionary,zhang2007learning}, with varying representations including HyperNEAT \cite{yosinski2011evolving}.  Control can be optimised on different levels, e.g. the gait level \cite{SoftwareEvolutionHexapodGait,HexapodWalkingGait}, or using a fixed gait and varying stride arcs and leg speeds \cite{7989128}.

Evolutionary Robotics aligns strongly with the philosophy of embodied intelligence, which states that body, brain, and environment must all be considered to elicit powerful in-environment behaviours (\cite{pfeifer_how_2006}, more recently~\cite{mengucc2017will}).  This raises an issue when evolving robots, in that both morphology and controller must be optimised, yet altering one often requires subsequent optimisation in the other~\cite{EvolvingRobotMorphologyAndControl,hornby_body-brain_2001,rosendo2017trade}.  Here, we compare three strategies for evolving body-brain couplings in our relatively constrained problem space; (i) using a `default' static controller, (ii) evolving controller parameters together with morphology parameters (single genome), and (iii) performing short-run controller optimisation for each morphology.  The notion is that constraining the permitted morphologies may permit the use of (ideally)(i), or (ii), which would reduce computational overheads.  Constraining the space in this way also makes it easier to tweak our chosen simulation environment towards being able to properly model these types of designs to improve future transfer to reality.

\subsection{Diversity-based evolution}
Quality-Diversity algorithms are a recent branch of evolutionary computing inspired by natural evolution~\cite{Markert2010}, which generate a library of potential solutions rather than using the population as genetic material to create a single fittest individual.  Because of their use of diversity, they are capable of producing higher-fitness solutions than optimising for fitness alone~\cite{pugh2016quality}.  

We focus on a particular quality-diversity algorithm, MAP-Elites~\cite{mouret2015illuminating}, which evolves a diverse library of local elites  -- that is, the algorithm fills out a library of individuals, replacing similar ones with fitter variants when they are found.  Similarity is compared using `features'; in the case of robotic morphology they could be related to physical aspects of the design, e.g., length of body, or number of legs.  As individuals are delineated based on features, they are an excellent choice for mapping out design spaces, and have been postulated as a key technology underpinning next-generation robotic design techniques~\cite{howard2019evolving}.

Robotics is a promising application area for Quality-Diversity algorithms, as evidenced by a number of recent works in the area applied to behaviour generation~\cite{mouret_encouraging_2012}, and combination with a probabilistic adaptation technique to provide fault recovery in legged robots~\cite{cully2015robots}.  For morphology generation, the automated discovery and optimisation of multiple morphological niches is demonstrated~\cite{lehman_evolving_2011}.  We differ from the above approach by applying domain knowledge in our encoding, restricting the types of solutions to those useful to our engineers, whilst also permitting more detailed exploration of viable brain-body couplings within those constraints.

To summarise, the principles of ER and diversity selection provide a powerful tool for exploration of complex design spaces, and are thus an excellent choice for implementing Design Assist for our large legged robots.

\section{Methodology}
\label{sec:methodology}

Direct and indirect encoding have both previously been successfully used to evolve robots~\cite{veenstra2017evolution,collins2018towards}; here we use a direct encoding of the robots morphology as it allows us to easily inject domain knowledge and constraints whilst allowing the exploration of a reasonably diverse set of morphologies.  Constraints are as follows:  maximum weight 60kg, forward speed >1m/s, minimum height 2m,  payload 10kg (battery, sensors, etc.).  Robots that fail any of these constraints are rejected and randomly regenerated.  We also investigate three controller implementations, in order of computational demand; 

\begin{itemize}
	\item STATIC: Each robot has the same, static, gait controller.
	\item GENOME:  Each robot has an evolvable gait controller, which forms part of the robot genome and is evolved along with it.
	\item ES: Each robot morphology runs a short 1+1 Evolution Strategy~\cite{beyer2002evolution} to optimise the controller.
\end{itemize}

\subsection{Morphology Genome}

Our morphology encoding defines the body and legs of the robot. The robot body is encoded by its cuboidal dimensions, and the horizontal placement of its centre of mass (in practice this is usually the location of payload and batteries). The remainder is compactly encoded by exploiting various symmetries: {\em bilateral symmetry} means we only define leg parameters for one side of the body, and these are mirrored to the other side.  {\em Link symmetry}, such that the thickness of the tube that represents each leg is the same across all links in that leg. 

\begin{figure}[h]
\centering
\begin{subfigure}[b]{.33\linewidth}
  \centering
  \includegraphics[width=.99\linewidth]{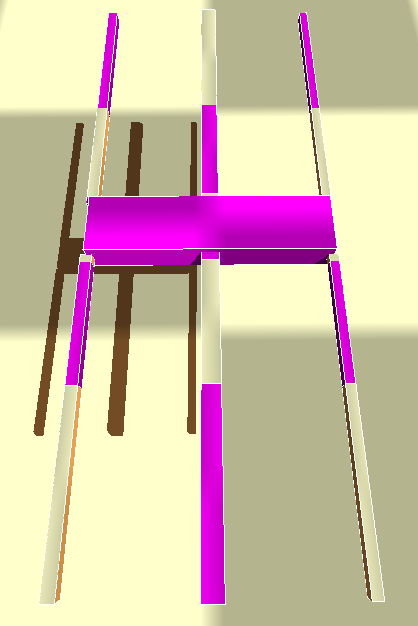}
    \caption{}
\end{subfigure}%
\begin{subfigure}[b]{.33\linewidth}
  \centering
  \includegraphics[width=.99\linewidth]{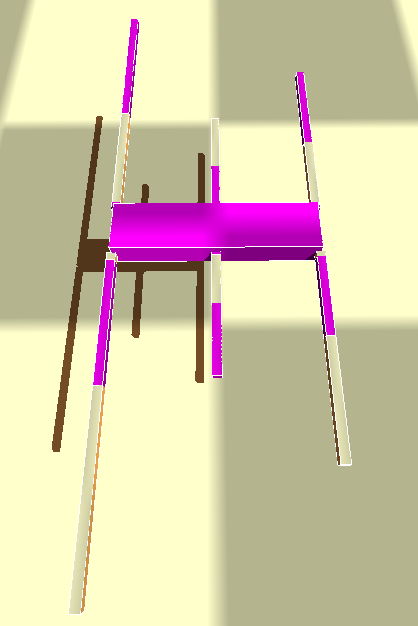}
    \caption{}
\end{subfigure}
\begin{subfigure}[b]{.33\linewidth}
  \centering
  \includegraphics[width=.99\linewidth]{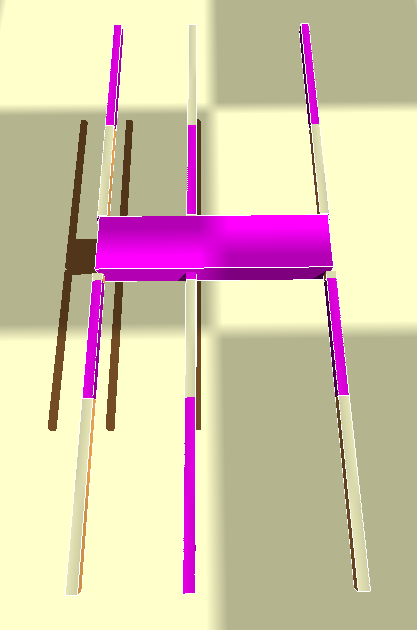}
    \caption{}
\end{subfigure}
\caption{Example of effects of quadratic scaling functions over the leg parameters. In order: the leg width scale (centre leg is wider), leg length scale (centre leg is shorter), and the attach location (centre leg attached left of centre) all follow quadratic functions (parabolas) with respect to the legs position index along the body. }
\label{fig:scales}
\end{figure}

Lastly, we include {\em leg symmetry} where parameters are the same for each leg, but with four quadratic variations. So we only encode one leg, which has three links, and repeat it down the body length. The per-link parameters (such as motor strength, damping and maximum torque) are equal for all legs. Four parameters\footnote{horizontal attach point onto the body, the leg's length and width scale, and a multiplier on its joint strengths} modify the overall leg parameters quadratically from front to back, such that e.g., front legs may be different sizes and have varying properties down the length of the robot, as per Figure~\ref{fig:scales}.  This compact encoding brings the search space from 222 scalars in the naive case of a 3-link hexapod down to 50 scalars regardless of the number of legs, as shown in Table~\ref{tab:morph_param}.   

\begin{table*}
  \caption{Morphology Genome Parameters}
  \label{tab:morph_param}
  \begin{tabular}{lcr}
    \toprule
    Type & Name & Range\\
    \midrule
    per robot & 3D body extents from centre & (0.4,0.025,0.05)--(1,0.15,0.3) m\\
              & 2D horizontal body centre of mass& (-0.5,-0.5)--(0.5,0.5) m\\
              & number of legs per side & 2--6\\
    per leg on a side & 3D attach point on body & (0,-1,0)--(0,1,0)\\
                      & leg attachment pitch angle & -0.6--0.6 rad\\
                      & leg tube thickness & 0.001--0.01 m\\ 
    per link on a leg & 3D cuboid extents & (0.0025,0.0025,0.0025)--(0.1,0.1,0.1) m\\
                      & 3D hinge axis & (0.8,-0.2,-0.2)--(1,0.2,0.2)\\
                      & & (-0.2,0.8,-0.2)--(0.2,1,0.2) on top link\\
                      & joint strength &2000--6000 Nm/rad\\
                      & joint damping &1--40 Nms/rad\\
                      & maximum joint torque &50--200 N\\
                      & maximum joint angular velocity &0--1.5 rad/s\\
    quadratic variation & 2D leg attach offset & linear: (0.7,-0.2)--(1,0.2) m\\
                        & & quadratic: (-0.1,-0.1)--(0.1,0.1) m\\
                        & leg length multiplier & -0.2--0.2\\
                        & leg width multiplier & -0.2--0.2\\
                        & leg strength multiplier & -5--5\\
  \bottomrule
\end{tabular}\\
3D offsets are (forward, up, right) in the robot frame. 2D offsets are (forward, right). 
\end{table*}

\subsection{Controller Genome}
Each robot is controlled using a parameterised sinusoidal gait generator, which controls desired joint angles and is guaranteed to be implementable in reality. We investigate three controller treatments as highlighted in Section~\ref{sec:methodology}.  

We assume a fixed gait pattern, the tripod gait, as a concession to genomic complexity.  This gait lifts the front and back legs from one side, and the middle leg from the other, per stride phase and is a good choice as it is statically stable.  For four and eight-legged robots, the pattern of lift-offs is maintained but commands are shortened or extended respectively.  We implement this gait by updating a single time parameter, and having each motor's sinusoidal output relative to this time using a single `stride' frequency $sf$, which is the first encoded parameter.

We encode the walking pattern efficiently by assuming bilateral and leg symmetry, leaving just one sinusoid to describe each link from body to foot. A general sinusoid of a given frequency has three free parameters ($a,b,c$): $y=a+b* \sin(sf* t + c)$, however we constrain the amplitude $b$ based on the maximum joint angular velocity $\omega$ encoded in the physical genome: $b=\omega_{max}/sf$. This leaves two parameters, the sinusoid vertical offset $a$ which we will now label $vo$, and its phase offset $c$ now labelled $po$. These two parameters are for each of the three links on a leg, together with the stride frequency $sf$, giving a compact genome with a total of seven scalars --- see Table~\ref{tab:controller} for details.  Default values for the STATIC controller are derived from MAX's default controller, shown in Figure~\ref{controlparams}(a).

\begin{table}
  \caption{Controller Genome Parameters}
  \label{tab:controller}
  \begin{tabular}{lcr}
    \toprule
    Type & Name & Range\\
    \midrule
    per robot & stride frequency $sf$ & 1--4 rad/s\\
    per link on a leg & sinusoid vertical offset $vo$ & -1--2 rad\\
                      & sinusoid phase offset $po$ & -$\pi$/2--$\pi$/2 rad\\
  \bottomrule
\end{tabular}
\end{table}

For STATIC experiments, controller values are set to the midpoints of each parameter, and do not change.  For GENOME experiments, controller values are random-uniformly set within range and are evolved when the rig is evolved, i.e., as part of the genome.  For ES experiments, controller values default to STATIC parameters, but for each rig a 1+1 Evolution Strategy is run for 20 iterations\footnote{For clarity we refer to robot evolution as running over generations, and controller evolution for ES as running over iterations.} to tune the controller to the robot morphology, for an approximate 20-fold increase in compute time.

\subsection{Simulation}
Experiments take place using the Bullet simulator\footnote{which is computationally fast and has an open API to allow us to flexibly embed domain knowledge of the problem to improve veracity.}, which is a commonly-used open source platform used extensively for robotics research. Dantzig's simplex algorithm is used to solve the multi-rigid-body dynamics as we have large bodies sitting above relatively low mass legs, which precludes the simpler iterative solvers as they do not converge efficiently in this problem domain.  

Robot joints are treated as driven linear spring-dampers with a maximum torque limit. The stiffness, damping and maximum torque are all evolved parameters.  The rigid parts of the robot are modelled as cuboids of evolved dimension parameters.  Because of this, the cuboids intersect whenever the joints have non-zero angle. We remedy this be turning off self-contact resolution between adjacent parts. However, more general self-intersections require some consideration. While the physics engine is capable of colliding legs together, there are two issues. Firstly, the physics engine does not consider the continuous motion of the legs for collisions, just their instantaneous position each simulation step, and for fast-moving thin legs, the engine may miss a collision altogether. Secondly, we do not want a robot that self-contacts during its walking motion. For these two reasons, we terminate a simulation run whenever a robot self intersection occurs (excluding those on adjacent parts). 

Robot mass is calculated empirically using MAX as a reference.  The body mass contains a fixed part of 7.5 kg representing the batteries and avionics. The remaining mass is treated as a uniform density of 170 kg/m$^3$.  Each leg part is treated as a hollow square cross-section tube, where the tube thickness is an evolved parameter and the density of the material is that of carbon fibre: 1600 kg/m$^3$. Additionally, the part contains the mass of the motor mechanism, which is assumed to be at the tip of the link. We model the mechanism as containing a fixed mass of 0.2 kg representing linkages and housing etc, plus the motor mass, which is correlated to the maximum power of the motor. The maximum power is the product of its maximum torque $\tau_{max}$ and maximum angular velocity output $\omega_{max}$, both evolved parameters. We derived the following equation empirically using data over a wide range of the popular Maxon servo motors used in robotics, see table~\ref{tab:maxon}.  Motor mass = $\sqrt{\tau_{max}\omega_{max}/13000}$

\begin{table}
  \caption{Maxon Motor Power and Mass}
  \label{tab:maxon}
  \begin{tabular}{lccr}
    \toprule
    Brand & Maximum power & Mass & W/kg$^2$\\
    \midrule
    DCmax16&4.3 W&0.023 kg &8129\\
    DCX14L&5.3 W&0.026 kg &7840\\
    ECX16L&107 W&0.073 kg &20079\\
    ECX19L&200 W&0.108 kg &17146\\
    ECX22L&233 W&0.148 kg &10637\\
    && mean: &12766\\
  \bottomrule
\end{tabular}\\

\end{table}

As Table~\ref{tab:maxon} shows, motors vary in power output by a factor of 44 and show a fairly constant power to square mass ratio (varying by a factor of 2.5).  An empirical model of motor mass is important because the legs themselves are assumed to be made of carbon fibre and are therefore very light.  Because of this, the mass of the joint motors is a significant contributor to the mass and inertia distribution of the robot. In addition to the adding realism, this provides a trade-off that prevents evolution from simply maximising maximum torque; a lighter robot might travel more efficiently, but must have less powerful motors to achieve the reduced weight.

\section{Evolutionary Procedure}
Robots are evolved following the MAP-Elites algorithm~\cite{mouret2015illuminating}.  Evolution parameters are given in Table~\ref{tab:params}, set after a brief parameter sweep.  At the start of an experiment, 400 individuals are created, assessed in simulation, and are assigned fitness $f$ and feature values $d1$-$d6$ before being placed into the 6D MAP-Elites library based on their calculated features.  ES individuals have their controllers optimised for 20 iterations.  Each individual either fills an empty bin, or replaces a current individual if its discretised feature dimensions match and it has a higher fitness.  We run 4000 generations, after which the experiment finishes.

\begin{table}
  \caption{Evolutionary rate settings}
  \label{tab:params}
  \begin{tabular}{lr}
    \toprule
    Name & Probability\\
    \midrule
    modify leg rate &0.25\\
    modify number of legs rate &0.25\\
    modify number of links rate &0.4\\
    modify motor rate&0.25\\
    modify leg offset rate&0.25\\
    modify body rate&0.25\\
    offspring per generation&60\\
    initial population size&400\\
  	\bottomrule
\end{tabular}
\end{table}

Every generation, 60 children are created by repeatedly selecting random individuals from the current library, and mutating following Table~\ref{tab:params}.  Mutations take place if a random sample from a uniform distribution in the range 0-1 is less than the relevant parameter value.  Mutations on real values add or subtract a value sampled from a normal distribution with covariance equal to 10$\%$ of the range of the parameter.  Mutations on integer values either add or subtract 1 from the current value.  All mutations are constrained to the permissible range of values for that parameter. See Table~\ref{tab:morph_param} for ranges.  Controller evolution for GENOME and ES affects every controller parameter, and follows the rules for real-valued parameter evolution.  For ES, the 1+1 ES replaces the current controller values with the child controller values at every iteration if the child is fitter than the parent. 

Feature dimensions are calculated and the individual is either assigned to a new bin or replaces a less-fit individual in an occupied bin as before.  The generation then ends.  We have 6 feature dimensions per robot, making this a rather high dimensional problem.  Each feature is discretised into 5 possible bins, for a total of 15625 feature combinations. The features, together with their ranges, are:

\begin{itemize}
\item d1: total leg length (before any scaling) [2.0,2.8] (m)
\item d2: total mass [50,75] (kg)
\item d3: number of legs per side [2,6]
\item d4: carbon fibre leg wall thickness [0.01,0.001] (m)
\item d5: leg length scale [-0.2,0.2]
\item d6: leg width scale [-0.2,0.2]
\end{itemize}

\section{Experimental Setup}

Each robot is subjected to a 5 second trial (enough to complete at least one gait cycle), which in this case involves the robot traversing a flat plane with the simulator running at 30Hz. The robots are fitness assessed according to their absolute cost of transport, which is $E/mgd$ for robot mass $m$, gravity $g$ and distance travelled in the forwards direction $d$. The expended energy is calculated as: $E=\int_{t=0}^5 \Sigma_i |\omega_i\tau_i| dt$, that is the time integral of the absolute mechanical power (angular velocity times torque) over all joints $i$.  To minimise this cost, the robot's fitness (which is maximised) is set to its reciprocal.

To assess statistical significance, we run 20 repeats of each experiment; STATIC, GENOME and ES, and perform Mann-Whitney U-tests with significance assessed at p$<$0.05.  Each experiment is run on a high performance compute cluster, with individual trials parallelised onto multiple cores on the node, and a single experimental repeat occupying one entire node.  ES experiments last approximately 3 days, STATIC and GENOME complete within 8 hours.

\section{Results \& Discussion}

 \begin{figure*}[t]
\begin{center}

\begin{subfigure}[b]{8.7cm}
  \centering
  \includegraphics[width=8.7cm]{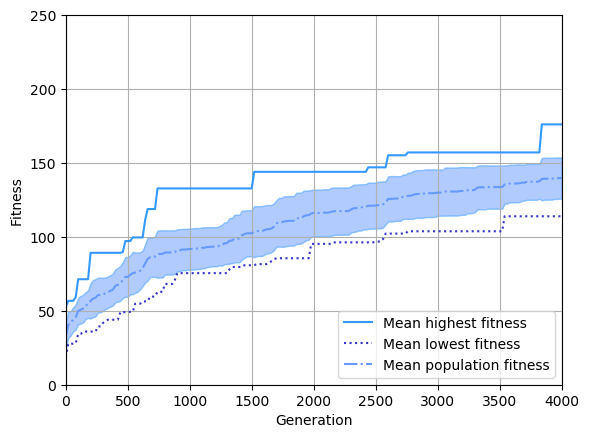}
  \caption{}
\end{subfigure}
\begin{subfigure}[b]{8.7cm}
  \centering
  \includegraphics[width=8.7cm]{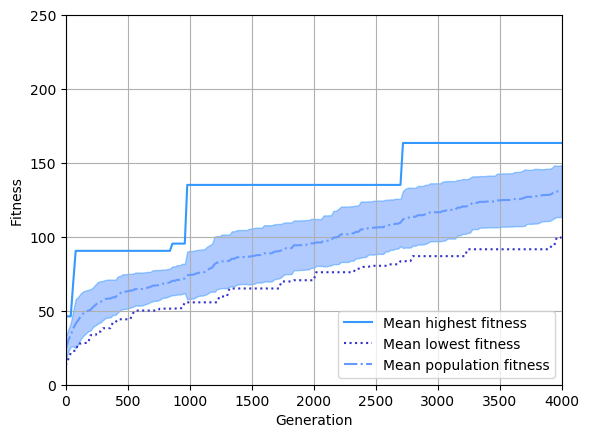}
  \caption{}
\end{subfigure}
\begin{subfigure}[b]{8.7cm}
  \centering
  \includegraphics[width=8.7cm]{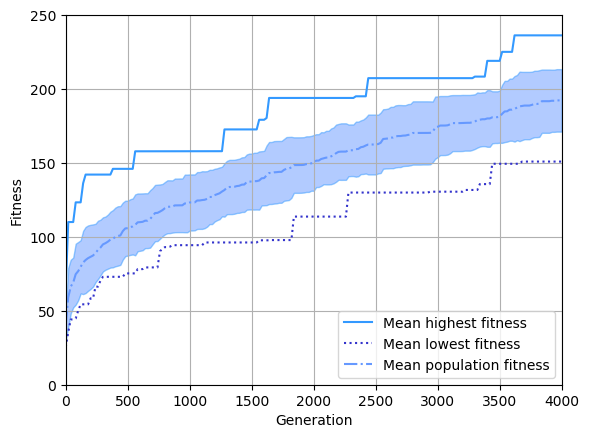}
  \caption{}
\end{subfigure}
\begin{subfigure}[b]{8.7cm}
  \centering
  \includegraphics[width=8.7cm]{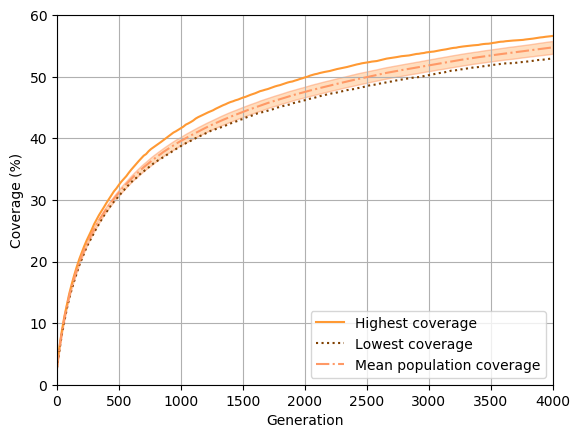}
  \caption{}
\end{subfigure}
\end{center}
\caption{Mean highest, average, and lowest fitness across all runs for (a) STATIC, (b) GENOME (c) ES.  ES displays significantly higher fitness than STATIC and GENOME. The shaded region denotes standard deviation from the mean. (d) Typical filling of the MAP-Elites library, averaged over all ES runs.  ES provides statistically superior coverage compared to GENOME, however all coverage progressions follow the same fill pattern.}
\label{Fitness}
\end{figure*}

Our chief performance indicator is fitness, or cost of transport. Averaged across the 20 repeats, we see that ES (Figure~\ref{Fitness}(a)) has the best high fitness  ($f$=192.32, s.d. 21.12), followed by STATIC (Figure~\ref{Fitness}(b)) ($f$=139.78, s.d. 13.94) and GENOME (Figure~\ref{Fitness}(c)) ($f$=130.97, s.d.17.5).  ES is significantly fitter than STATIC and GENOME (p$<$0.05), due to the extended computational effort expended in optimising the controller.  What is perhaps unexpected is that the GENOME scheme does not outperform the STATIC default walk controller. An interpretation of this result is that simultaneous morphology and controller evolution makes the optimisation task harder due to the larger search space, and the constant mutation of the walk controller makes it hard for the morphology parameters to converge.

\subsection{Coverage}

Figure~\ref{Fitness}(d) shows a typical filling of the MAP-Elites library through the generations, with the expected sharp rise over the first $\approx$ 500 generations followed by a gradual plateau to average final coverage values of STATIC = 54.7$\%$, GENOME = 53.79$\%$, ES=54.76$\%$.  Standard deviations are 0.73, 0.74, and 1.02, respectively.  The interesting result is that {\em optimising the controller with ES allows more of the design space to be accessed}, with mean coverage for ES being significantly higher than GENOME (p$<$0.05).  Coverage values seem a little low, but given the difficulty in simulating this class of robot, with e.g. the potential for leg collisions, is reasonable.  

\begin{figure*}[t]
\begin{center}
\begin{subfigure}[]{10cm}
  \centering
  \includegraphics[width=10cm]{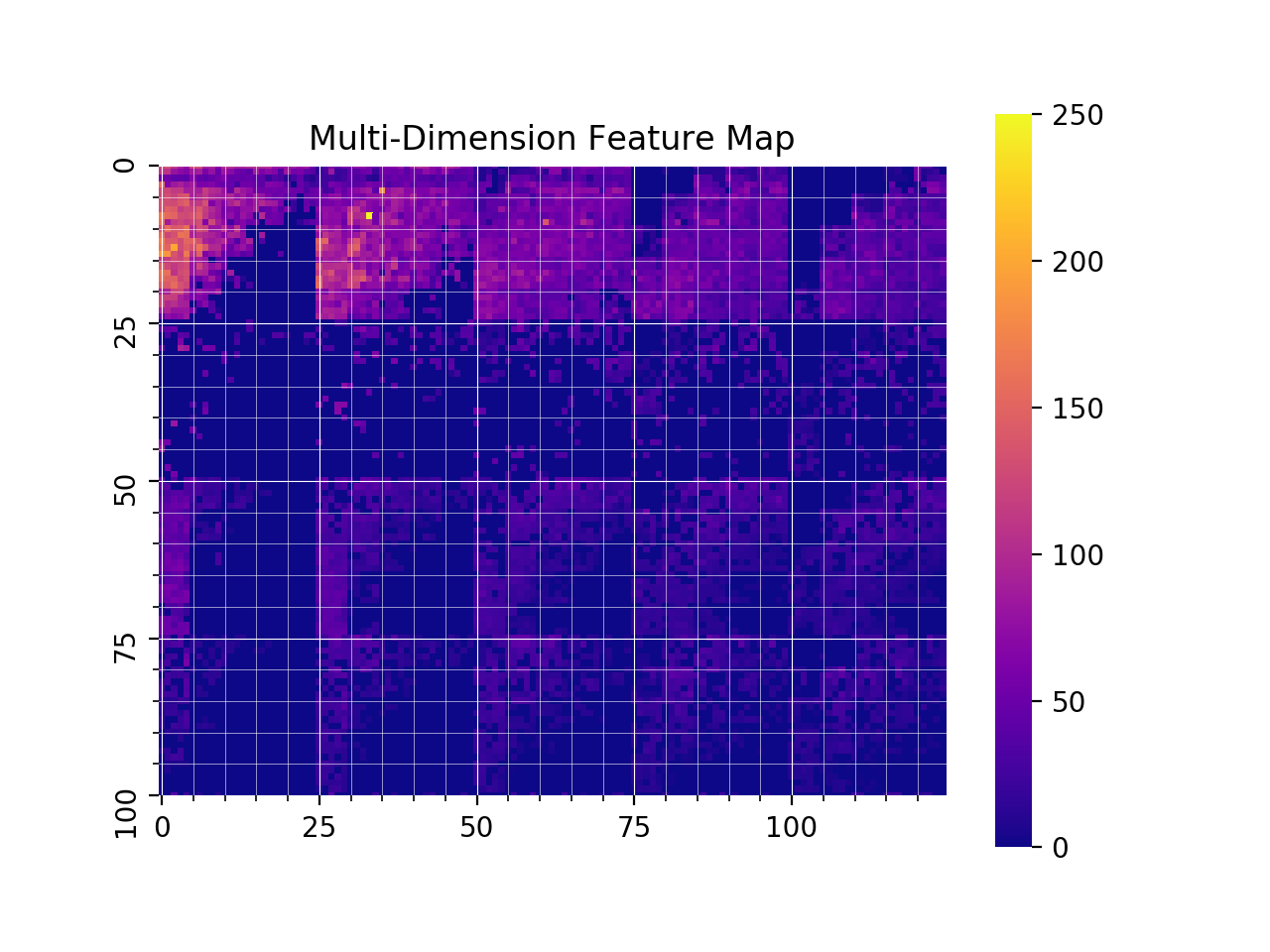}
\end{subfigure}
\begin{subfigure}[]{6cm}
  \centering
  \includegraphics[width=6cm]{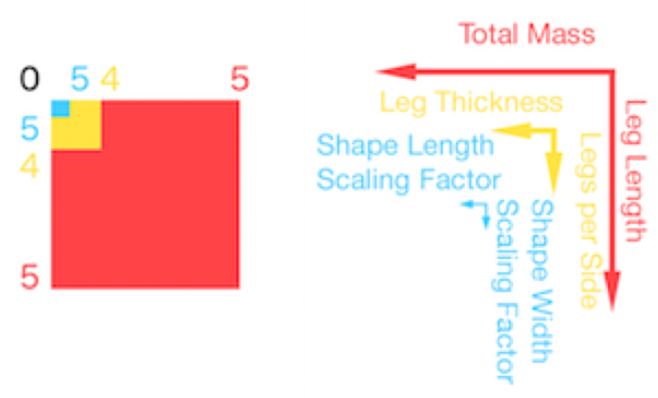}
\end{subfigure}
\end{center}
\caption{Sample feature map for a run of ES.  Feasible regions of the design space as well delineated.  The original MAX has been added as the white point, sitting close to a fitness `hot spot'.}
\label{mapppp}
\end{figure*}

An exemplar library is given Figure~\ref{mapppp}.  Given the regular shapes of the fitness hot spots in the final feature map, and our use of direct encoding to ensure that all areas of the space can be uniformly sampled, we can be fairly sure on the regions of feasible design space we can operate in.  The original MAX has been added as the bright white spot, sitting close to, but not exactly on, a hot spot.  This goes some way to showing the veracity of our simulation approach, whilst also offering our engineers various potential other parametric design combinations to consider.  

\subsection{Controller Optimisation}

\begin{figure*}[t]
\begin{center}

\begin{subfigure}[b]{5.7cm}
  \centering
  \includegraphics[width=5.7cm]{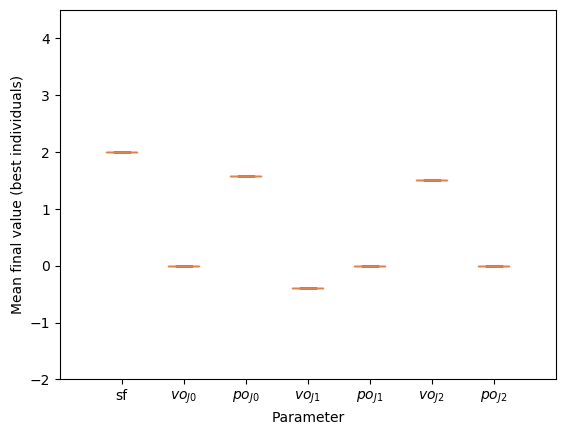}
    \caption{}
\end{subfigure}
\begin{subfigure}[b]{5.7cm}
  \centering
  \includegraphics[width=5.7cm]{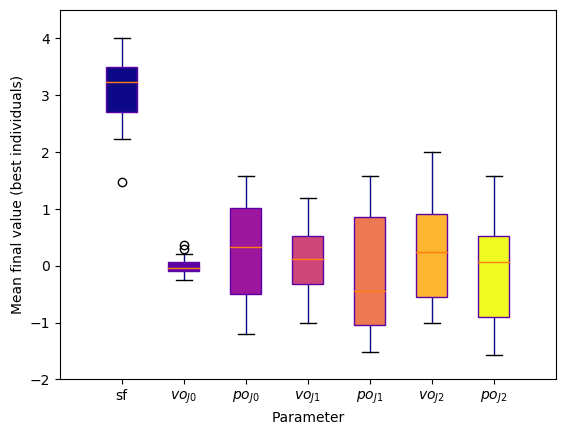}
    \caption{}
\end{subfigure}
\begin{subfigure}[b]{5.6cm}
  \centering
  \includegraphics[width=5.6cm]{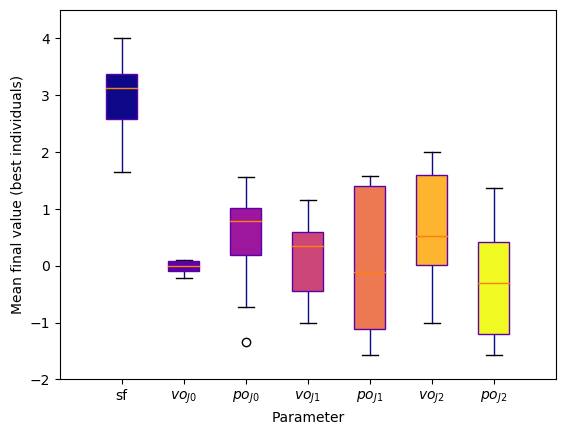}
    \caption{}
\end{subfigure}
\end{center}
\caption{Controller parameters for (a) STATIC, (b) GENOME, and (c) ES, averages over the best individual per repeat.  Although distributions are similar for (b) and (c), the ability to iteratively tailor control to morphology provides heightened fitness for ES.}
\label{controlparams}
\end{figure*}

An aim of this paper is to explore the extent that body-brain couplings matter in relatively constrained design problems.  Mean controller values across the single best individual in each repeat are shown in Figure~\ref{controlparams}.  In terms of informing the design of MAX, we also note that the stride frequency is indicated by GENOME and ES to be feasible at $sf$=3, rather than $sf$=2 as per the default controller.  Other than that, phase offset for joint 0 `$po_{J0}$' is indicated to be higher in the static controller than is desired.  Overall, ES optimised parameters follow reasonably the STATIC parameters, and are visibly similar to GENOME parameters, therefore we conclude that the ability to tailor control specifically to a given morphology is critical, even in constrained problems such as ours.

\subsection{Designs}

\begin{table}
  \caption{Mean parameter values for top 10\% robot designs}
  \label{tab:results}
  \begin{tabular}{lccc}
    \toprule
    Grid parameter & STATIC & GENOME & ES\\
    \midrule
    $\bar{d}$1: leg length & 2.31 m & 2.34 m & 2.27 m\\ 
    $\bar{d}$2: mass & 65.4 kg & 65.9 kg & 61.8 kg\\ 
    $\bar{d}$3: number of legs per side & 4.0 & 3.9 & 4.0\\ 
    $\bar{d}$4: leg wall thickness & 3.4 mm & 3.6 mm & 3.4 mm\\ 
    $\bar{d}$5: leg length scale & -0.01 & 0.003 & -0.003\\ 
    $\bar{d}$6: leg width scale & 0.002 & -0.0006 & -0.002\\ 
  	\bottomrule
\end{tabular}
\end{table}

\begin{figure*}[h]
\centering
\begin{subfigure}[b]{.5\linewidth}
  \centering
  \includegraphics[width=.99\linewidth]{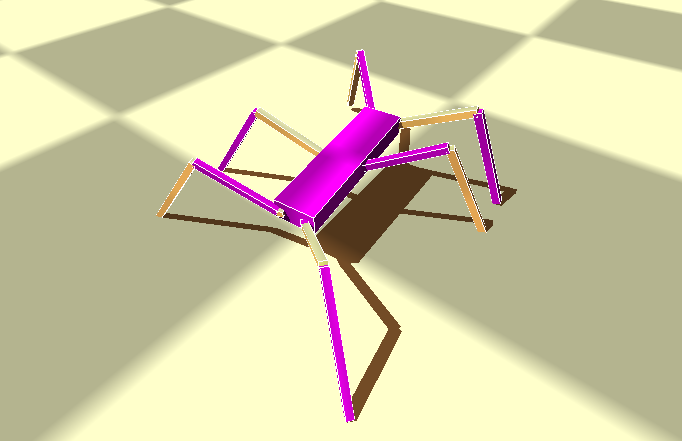}
  \caption{}
\end{subfigure}%
\begin{subfigure}[b]{.5\linewidth}
  \centering
  \includegraphics[width=.99\linewidth]{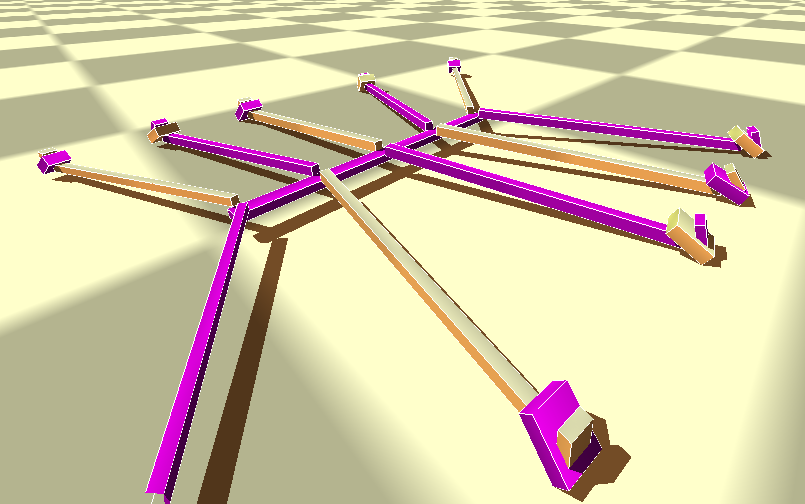}
  \caption{}
\end{subfigure}
\caption{Example evolved robot designs. Left: A typical high-fitness walker design. Right: A low profile design, with a low energy (but high structural stress) `pond skater' walking strategy. }
\label{fig:exampledesigns}
\end{figure*}

The results in table~\ref{tab:results} show typical parameter values for the 90$^{th}$ percentile across STATIC, GENOME and ES, that the main parameters vary little with the controller evolution scheme. However, the 90$^{th}$ percentile fitness is quite different between the three types. We interpret this as due to the different evolution schemes producing different quality walk patterns for morphologies.  Typical robots are presented in Figure~\ref{fig:exampledesigns}. 6(b) in particular may not be useful (unless seeking to skate over ice!) and are easily removed from the design study.

\section{Generating design rules}

\begin{table*}[ht!]
  \caption{ES design rules.  Please refer to main text for explanation of the rules.}
  \label{tab:designrules}
  \begin{tabular}{lc}
    \toprule
    Rule & Mean error \\
    \midrule
leg length = $\bar{d}1 -47.1(d4 - \bar{d}4) -0.0339(d3 - \bar{d}3) + 0.000214(d2 - \bar{d}2) + 0.0644(d6 - \bar{d}6) -0.0596(d5 - \bar{d}5)$ & 9.09\% \\
tube thickness = $\bar{d}4 -0.00139(d3 - \bar{d}3) + 7.71e-05(d2 - \bar{d}2) + 0.000761(d1 - \bar{d}1) -0.000408(d5 - \bar{d}5) + 0.000238(d6 - \bar{d}6)$ & 18\% \\
  	\bottomrule
\end{tabular}
\end{table*}

As part of our Design Assist methodology, we attempt to approximate the results of the ES experiments as a set of robot design rules. An example of such a rule would be that a hexapod with low cost of transport requires its legs to be a particular multiple of its body length. In general this is a set of rules that approximate the optimal value of one design parameter as a function of the other parameters, which is a useful tool for navigating unintuitive and complex design spaces.  For simplicity we only look at linear functions in this paper, and therefore the basis of these design equations is a Single Value Decomposition. 

Taking the top 10\% highest fitness individuals over the 20 repeats, we refer to each robot design by the six features that are used by MAP-Elites, giving a six-dimensional vector valued genome $\vec{g}_i$ for indices $i=1..n$. We can then form the covariance matrix as $M=\Sigma_i (\vec{g}_i-\bar{\vec{g}})(\vec{g}_i-\bar{\vec{g}})^\top/n$ where $\bar{\vec{g}}$ is the mean genome over all $i$. This covariance matrix represents a six-dimensional ellipsoidal distribution of successful robot designs, where the thinnest axes represents the highest correlations.  To minimise bias resulting from unequal parameter ranges, we approximate all parameters as linear quantities centred at zero, and define their overall scales as root mean square values:

\begin{equation}
\vec{s}_j=\sqrt{\frac{\Sigma_j \vec{g}_{ij}^2}{n}}
\end{equation}

We normalize the covariance matrix by dividing the mean-relative vector $\vec{g}_i-\bar{\vec{g}}$ elementwise by the scale factors $s_j$ (where $j$ indexes each parameter in the genome) and subsequently perform an eigenvalue decomposition to convert the covariance matrix into a vector of eigenvalues $\vec{e}$ and a matrix of eigenvectors $V$, such that $M=VI\vec{e}V^\top$. These represent the extents and orientation of the covariance of successful robot designs. 

We retain the most significant correlations, reject any $i$ once $\sqrt{\vec{e}_i}>0.2$, and present each as a direct formula, in order of significance.  This produces a list of design equations of decreasing significance, where each equation has terms of decreasing importance from left to right. This allows the robot designer to trade off design accuracy with simplicity, and to gain insight into which parameters, and which parameter relationships, matter. The rules for the best performing (ES) evolution scheme are shown in table~\ref{tab:designrules}.

The design rules in table~\ref{tab:designrules} show that there are only two linear rules with less than the threshold of 20\% error, and of these, the first rule is twice as precise as the second. If we look at just the first three terms one can see that the robot leg length is negatively correlated with it's tube thickness, negatively correlated with the number of legs and positively correlated with the total robot mass.  This method of analysis appears effective in simplifying a large amount of data from the MAP-Elites library into a useful set of guidelines for designers.

\section{Conclusions \& Outlook}

We have shown how MAP-Elites can be used as a Design Assist tool to shed light on an under-studied design space of large legged robots.  There are a number of important takeaways from this work:
In comparing three control evolution strategies, we have shown that body-brain couplings must be prioritised, even in morphologically-constrained problem spaces. This does not mean simply allowing the controller to evolve, but that adequate compute time must be given to allow good couplings to emerge.
Controller evolution via ES is not just a tool for encouraging high performance solutions; controller optimisation in our case opened up significantly more of the problem space in terms of solution coverage when compared to GENOME.
As a design tool, we can see how MAP-Elites helps to map out the feasible regions of the design space.  Moreover, we have demonstrated how a small set of useful, general rules can be pulled from the MAP-Elites library to inform designers operating in that space. We consider this ordering of rules and terms as an efficient way to present the results to robot design engineers.

\bibliographystyle{ACM-Reference-Format}
\bibliography{gecco-largehex} 

\end{document}